\documentclass[10pt,leqno]{amsart}

\usepackage{graphicx}
\usepackage{indentfirst,csquotes}
\usepackage{amssymb,amsthm,amsmath}
\usepackage{xcolor,paralist,hyperref,fancyhdr,etoolbox}

\usepackage{subcaption}
\usepackage{multirow}
\usepackage{booktabs}
\usepackage{float}

\hypersetup{ 
    colorlinks=true, 
    linkcolor=black, 
    filecolor=black, 
    urlcolor=black,
    citecolor=black
}

\begin{document}

\title{Why Evolve When You Can Adapt? Post-Evolution Adaptation of Genetic Memory for On-the-Fly Control}

\author[H. Hammami]{Hamze Hammami}
\address{School of Engineering and Physical Sciences, Heriot-Watt University Dubai}
\email{hh2095@hw.ac.uk}

\author[E. D. Barbulescu]{Eva Denisa Barbulescu}
\address{School of Engineering and Physical Sciences, Heriot-Watt University Dubai}
\email{edb2000@hw.ac.uk}

\author[T. Shaikh]{Talal Shaikh}
\address{School of Mathematical and Computer Sciences, Heriot-Watt University Dubai}
\email{t.a.g.shaikh@hw.ac.uk}

\author[M. Aldada]{Mouayad Aldada}
\address{School of Engineering and Physical Sciences, Heriot-Watt University Dubai}
\email{ma3012@hw.ac.uk}

\author[M. S. Munawar]{Muhammad Saad Munawar}
\address{School of Engineering and Physical Sciences, Heriot-Watt University Dubai}
\email{mm4044@hw.ac.uk}

\date{\today}

\maketitle

\begin{center}
\textit{Accepted for publication at ALIFE 2025: The 2025 Conference on Artificial Life}\\
\textit{International Society for Artificial Life (ISAL) $\cdot$ \url{https://2025.alife.org/}}
\end{center}
\bigskip

\let\thefootnote\relax

\begin{abstract}
Imagine a robot controller with the ability to adapt like human synapses, dynamically rewiring itself to overcome unforeseen challenges in real time. This paper proposes a novel zero-shot adaptation mechanism for evolutionary robotics, merging a standard Genetic Algorithm (GA) controller with online Hebbian plasticity, drawing inspiration from biological systems through the separation of learning and memory, with the genotype acting as memory for the robot and the learning through Hebbian updates. In our approach, the fitness function itself is leveraged as a live scaling factor for Hebbian learning, so that the robot's neural controller can adjust its synaptic weights on-the-fly without any additional training. This adds a dynamic adaptive layer that activates only during runtime to handle unexpected changes to the environment. After the task, the robot "forgets" the temporary adjustments and reverts to the original weights, preserving core knowledge. We validate this hybrid GA–Hebbian controller on an e-puck robot in a T-maze navigation task with changing light conditions and obstacles.
\end{abstract}

\section{Introduction}
Genetic algorithms (GA) controllers excel in static or predictable environments, where a fixed control policy optimized during evolution can perform well.Genetic Algorithms (GA) are defined through a process of evolution inspired by Darwin’s theory of natural selection. Following that However, the evolution process itself adapts to static environments, or multiple environments when given. As Mori and Kita observed, "Though GA show good performance in solving a static optimization problem, GA sometimes fail to adapt to a dynamic environment" \cite{mori2000_ga_dynamic}.
Even minor environmental changes can cause a controller evolved for one scenario to fail, requiring costly and time-consuming retraining from scratch to adapt to a single change.
Several approaches have been explored to introduce online adaptation in evolutionary controllers. For example, \cite{5533882} introduces GPIS, an intelligent system for control using a "Percepter' and an "rAdaptor" as a Symbolic Rule Controller. This system continuously monitors whether the current IF–THEN rule is handling the environment using the Percepter. The rAdaptor then monitors the Percepter, and when it flags "inadaptable," it triggers a flexible genetic programming (fGP) at runtime to evolve and replace the control rule on-the-fly.
A more bio-inspired route is to leverage principles of neural plasticity. Hebbian learning, often summarized as "neurons that fire together, wire together" (Hebb, 1949), provides a simple local rule for adjusting connection strengths based on co-activation. However, Hebbian plasticity alone can be unstable if unregulated, as stated by \cite{ZENKE2017166}. Synaptic weights might grow without bound or decay to zero, causing unstable behavior if left unchecked and unmonitored. Prior studies have addressed this by introducing compensatory mechanisms. A notable expansion is Spike‐Timing‐Dependent Plasticity (STDP) \cite{Caporale2008}. This approach solves the problem by implementing 'potentiation' and 'depression' to the precise timing of spikes, introducing bidirectional weight adjustments contingent upon the temporal sequence of pre-synaptic and post-synaptic neural activation. When pre-synaptic spikes precede post-synaptic firing within a critical time window, the connection undergoes Long-Term Potentiation (LTP); conversely, if post-synaptic activity precedes pre-synaptic input, Long-Term Depression (LTD) occurs, systematically weakening the synaptic efficacy. Temporal correlation-based plasticity introduces intrinsic homeostatic regulation that conventional Hebbian mechanisms lack.
Both STDP and Hebbian learning mimic the concept of plasticity, but plasticity isn't necessarily applied only in these algorithms. A recent research direction relocates the plasticity concept from neural connections within individual controllers to the interconnections between agents in a multi-agent system, \cite{10.1162/isal_a_00799} demonstrate this through an enhanced Boids model where each pair of flock members maintains a mutable social-influence weight that undergoes periodic updates based on transfer-entropy measurements. Agents that consistently influence the movement of others experience strengthening of their outgoing connection weights, while those with minimal influence undergo connection attenuation. This mechanism facilitates the spontaneous reorganization of the swarm into diverse topological formations including chains, rings, and emergent leader-follower patterns without modifying the underlying movement rules that govern individual behavior.
Approaches towards Hebbian hybrids have a significant impact in improving genetic algorithms; mainly, our proposed methodology aims to improve post-evolution behavior and will provide a comparative review to understand the differences and the aims of Hebbian and genetic approaches.
An approach to meta-learning Hebbian plasticity \cite{NEURIPS2020_ee23e7ad} aims to self-organize randomly initialized neural networks through connection-specific plasticity rules (ABCD tables). This mechanism enables a random network to self-organize from scratch and repeatedly re-learn throughout each episode. The evolution process runs on the plasticity rules themselves, with the primary goal of improving quadruped locomotion and damage recovery. \cite{Yaman_2021} uses a similar approach by encoding plasticity rules in a finite search space and optimizing them with genetic algorithms with the primary goal of adapting to changing environmental conditions for 'Autonomous Learning'.
\cite{10.1145/3449639.3459317} initially assign one ABCD rule per synapse, then systematically cluster and evolve these rules to minimize redundancy. Their method transitions from per-synapse initialization to cluster-shared rule subsets (approximately 1,900 parameters after merging), allowing continuous weight updates while maintaining adaptive capabilities. \cite{10.1145/3638529.3654011} present a neuron-centric Hebbian learning approach that significantly reduces parameter complexity by assigning one ABCD rule per neuron rather than per synapse and initializes weights via a separate genetic algorithm, then continuously updates all inbound weights of a neuron using that neuron's specific rule.

The approach we propose is a manifestation of the Baldwin effect \cite{sznajder2012adaptive}. It is the approach where phenotypic adaptability or learning during an individual’s lifetime is not directly inherited by its offspring, but can influence evolutionary success over time. We implement this idea through a two-phase mechanism, GA evolve and optimize neural controller's baseline weights encoding the foundational "genetic memory",then we apply Hebbian adaptation learning during the post evolution phase which enables the offspring controllers to dynamically adjust its interaction during its runtime.In practice, this means each robot begins its operational lifetime with fixed neural parameters, and then independently undergoes Hebbian adaptation when interacting with its environment.

\section{Methodology}
This research introduces a post-evolution approach to implementing neuro-plastic adaptation mechanism, initially, we utilize a static environment train the robot in through GA, this is an isolated process, and a fitness function is employed for performance evaluation, then we apply plasticity through the Hebbian rule to adapt the 'genetic memory' with fitness being outlined as performance feedback.
\subsection{Robot and environment}
\begin{figure}[ht]
\centering
\includegraphics[width=0.6\textwidth]{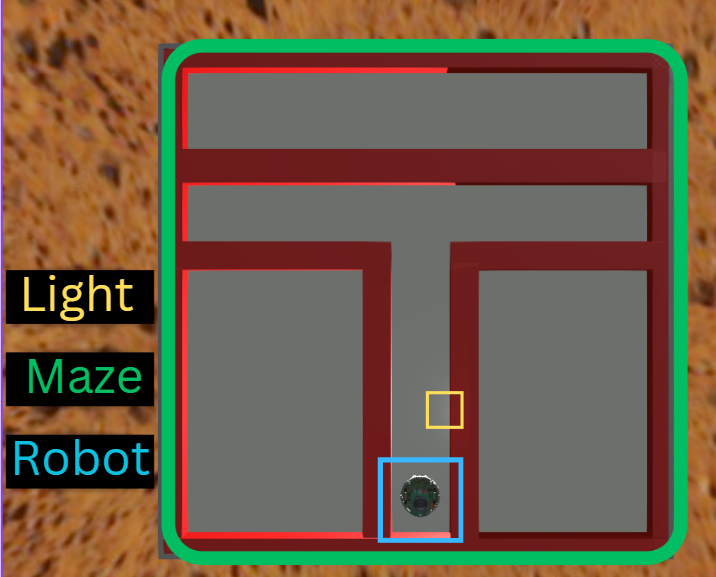}
\caption{A base robot environment}
\label{fig}
\end{figure}

We validate our methodology using the e-puck robot in the Webots simulation platform. The e-puck is a small wheeled robot equipped with multiple sensors; in our setup we utilize its eight infrared proximity sensors and eight light sensors. The test environment is a T-shaped maze, which can be seen in \textbf{Fig:\ref{fig}}.
\textbf{Fixed elements:} These are aspects of the environment that remain constant during the evolutionary training phase. the overall maze layout and two sets of goal locations which depends on light configurations. One with light and one without light. The structure is a T-shaped maze in which the robot's navigation direction is determined by the light sensor's readings and the presence of a specific light source. The tasks in this environment are: if the light is present, the robot should turn right at the junction; if light is absent, it should turn left. The threshold is not fixed and depends on the controllers and the environment it was trained on. The second fixed element category pertains to world settings, which define the training environment, such as background illumination which can affect the light sensor readings, background light level is fixed only during evolution.

\textbf{Dynamic elements:} These refer to changes introduced during testing to evaluate the controller’s adaptability. We consider two types of perturbations, first a light intensity change where after evolving the robot in a brightly lit scenario, we drastically dim the light during testing, second introducing obstacles – we place additional wall segments in the corridors, altering their width. We design these dynamic changes to not make the task impossible but to require adaptation, and goal point changes are currently excluded as potential dynamic elements but can be implmented in further work. 
This environment is currently used to benchmark, propose and further develop the methodology. 

\subsection{GA and Fitness}
This methodology implements a base evolutionary robotics (ER) approach as the first stage. This is necessary in order to establish a base controller. Using this base knowledge, we intentionally give the robot limited environments without the environment changes needed to test the adaptability in the second phase. The ER network is implemented on a Multi-Layer Perceptron (MLP) controller, with its weights serving as the genotype. The MLP has 16 input neurons (8 light sensors + 8 distance sensors), 4 hidden layers with [7,5,8,4] neurons, and 2 output neurons for motor control, using a supervisor-controller framework where the supervisor manages robot data and oversees training and trials, while the controller processes sensor inputs to generate motor commands.

The genotype, represented as a real-valued vector of MLP weights, is optimized using a genetic algorithm that involves population initialization with random weights, fitness evaluation, selection for reproduction, crossover to combine genotypes, and mutation to introduce diversity. The evolutionary parameters used are 30 generations, population size of 50, and elitism of 6.

A dual fitness function system combines "behavior fitness" and "goal reward" to ensure the robot exhibits desired behaviors. The behavior fitness components are defined as:
\begin{equation}
\label{eq:fitness-components}
\begin{aligned}
\text{forwardFitness} &= \frac{v_{\text{left}} + v_{\text{right}}}{1.5}, \\[6pt]
\text{avoidCollisionFitness} &= 1 - 
\bigl(\max(\text{proximitySensors})\bigr)^{3}, \\[6pt]
\text{spinningFitness} &= 1 - \frac{\bigl|v_{\text{right}} 
- v_{\text{left}}\bigr|}{2}, \\[6pt]
\text{junctionFitness} &=
\begin{cases}
1, & \text{if correct turn is made},\\
0, & \text{otherwise}.
\end{cases}
\end{aligned}
\end{equation}

These components are combined with priority given to collision avoidance:
\begin{equation}
\mathrm{combinedFitness}
= \frac{f_{\mathrm{forw}} \;+\; 2\,f_{\mathrm{avoid}} \;+\; f_{\mathrm{spin}} \;+\; f_{\mathrm{junction}}}{5}.
\end{equation}

The goal reward is based on Euclidean distance to the target position:
\begin{equation}
\label{eq:reward}
\begin{aligned}
\text{dist} &= \sqrt{(x_{g} - x)^{2} + (z_{g} - z)^{2}}, \\[6pt]
\text{distScaled} &= \min(1, (1.7 \times \text{dist})^3), \\[6pt]
\text{reward} &= 1 - \text{distScaled}.
\end{aligned}
\end{equation}

The final fitness value is computed as the average of the behavior fitness and goal reward:
\begin{equation}
\text{finalFitness} = \frac{\text{combinedFitness} + \text{reward}}{2}
\end{equation}
This fitness function is tailored to our T-shaped maze, with feedback provided based directly on the robot's desired performance that we need. 

\subsection{Post-evolution Hebbian}
The proposed methodology introduces a post-evolution adaptation process aimed at enhancing the adaptability of evolved controllers to dynamic environments. While the base genotype, optimized through evolutionary algorithms, demonstrates robust performance in fixed training environments and due to limited environment changes we explicitly set, it often encounters challenges when deployed in scenarios with these environment changes. To overcome this limitation, we incorporate principles inspired by Hebbian learning, a foundational mechanism of synaptic plasticity in biological neural networks. As Donald Hebb famously stated in his seminal work \cite{Hebb1949}:

\begin{quote}
"When an axon of cell A is near enough to excite a cell B and repeatedly or persistently takes part in firing it, some growth process or metabolic change takes place in one or both cells such that A's efficiency, as one of the cells firing B, is increased."
\end{quote}
In other words, the Hebbian rule posits that the strength of the connection between two neurons increases when they are consistently activated together. 
\begin{figure}[ht]
  \centering
  \includegraphics[width=0.8\textwidth]{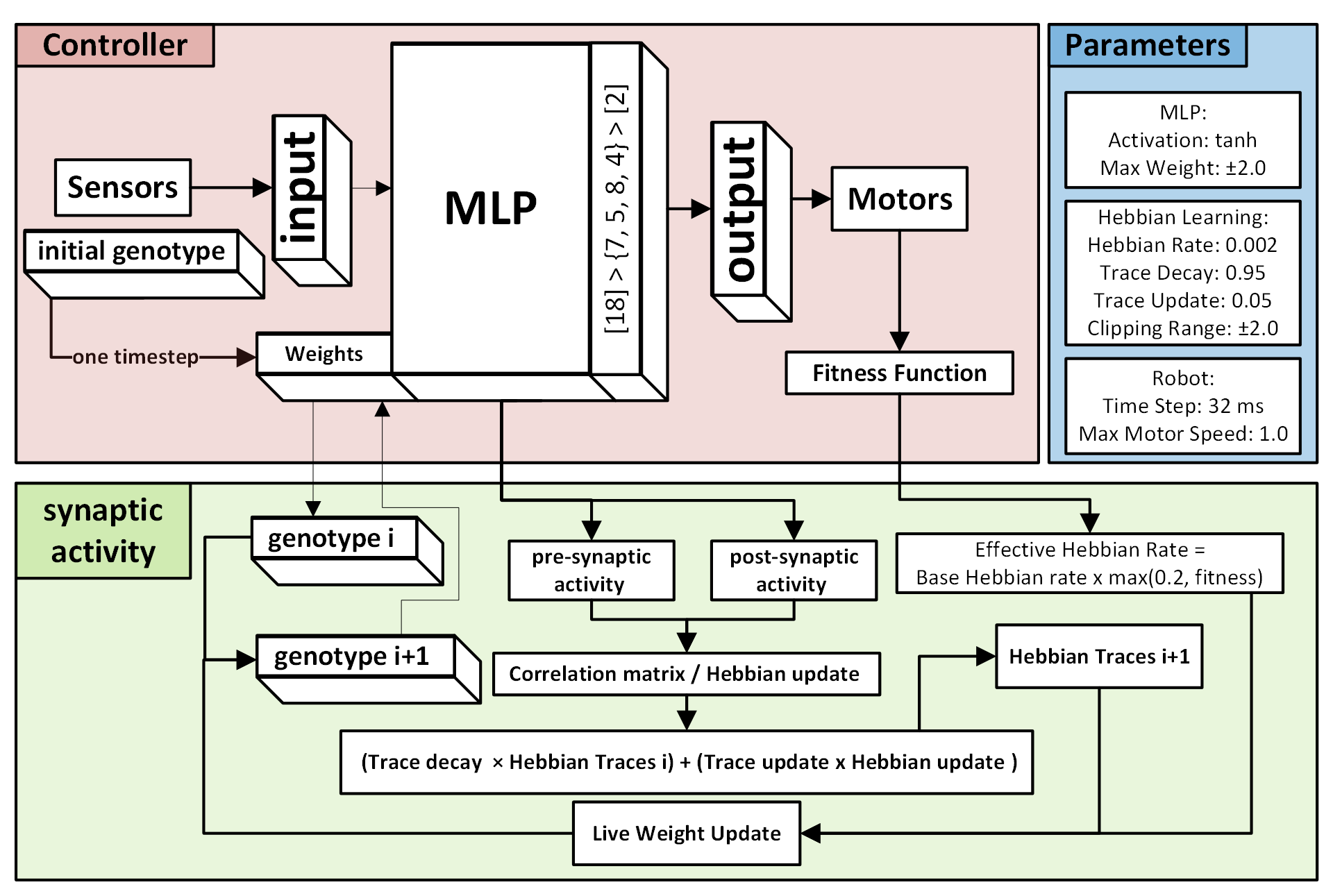}
  \caption{Hebbian gene adaptation workflow}
  \label{fig:Hebbian arch}
\end{figure}

The demonstrated workflow in \textbf{Fig:\ref{fig:Hebbian arch}} illustrates the operation of the controller and how synaptic activity is computed, showing the communication between both components. In this configuration, the controller's genotype remains static when running the MLP controller alone, showing no adaptation. By incorporating the Hebbian approach and using the previously defined fitness function as feedback, we can determine an effective learning rate through live fitness evaluation, which helps assess the robot's current motor performance. This live fitness is applied to the Hebbian rate, serving as a scaling factor to adjust the connection weights in real time.

\noindent Hebbian Rate formula: 
\begin{equation}
N_e = N \times \max(0.2, F)
\end{equation}

\noindent where:
\begin{itemize}
    \item \(N_e\): effective Hebbian rate applied to weight updates
    \item \(N\): the base Hebbian rate parameter
    \item \(F\): current fitness value
\end{itemize}

\noindent This results in a learning rate proportional to the robot's performance, a form of neuromodulation signal. 
example possible scenarios such as:

\noindent \textbf{Low Fitness Scenario:}\\
When the robot performs poorly with $F = 0.1$:
\begin{equation*}
\begin{aligned}
\max(0.2, 0.1) &= 0.2 \\
N_e &= 0.002 \times 0.2 \\
N_e &= 0.0004
\end{aligned}
\end{equation*}
\noindent The effective learning rate becomes 20\% of the base rate, limiting weight adjustments during poor performance while still keeping a minimal adaptation for exploration.

\noindent \textbf{High Fitness Scenario:}\\
When the robot performs well with $F = 0.9$:
\begin{equation*}
\begin{aligned}
\max(0.2, 0.9) &= 0.9 \\
N_e &= 0.002 \times 0.9 \\
N_e &= 0.0018
\end{aligned}
\end{equation*}

\noindent The effective learning rate reaches 90\% of the base rate, allowing robust reinforcement of successful neural pathways during high-quality behaviors.

\noindent Essentially, the better the performance, the faster the Hebbian mechanism will strengthen connections between active neurons. Higher fitness values accelerate the reinforcement of successful neural pathways, while lower performance results in slower learning rates to prevent erratic behaviors during periods of poor adaptation.

We implement a trace mechanism with weight decay:
\begin{equation}
T_{i+1} = (T_{\text{decay}} \times T_i) + (T_{\text{update}} \times C)
\end{equation}
where:
\begin{itemize}
    \item \(T_i\): Old Hebbian trace
    \item \(T_{\text{decay}}\): Decay factor reducing the influence of older updates
    \item \(T_{\text{update}}\): Integration factor controlling the influence of the current Hebbian update
    \item \(C\): Correlation matrix
\end{itemize}
This trace mechanism is essential to address a limitation of Hebbian learning discussed earlier in \cite{ZENKE2017166}, its instability due to positive feedback loops. The trace decay factor is set to 0.95, meaning that 95\% of the previous trace information is retained while 5\% is replaced with new correlation data. This creates a temporally-extended record of neural co-activations, preventing rapid weight oscillations and providing resilience against momentary spurious correlations.
\newpage

After calculating the effective Hebbian rate, we use the traces along with this rate to determine the weight changes and apply them to establish a new genotype:
\begin{equation}
\Delta W = N_e \times T_{i+1}
\end{equation}
\begin{equation}
W_{T+1} = W_T + \Delta W
\end{equation}
where:
\begin{itemize}
    \item \(\Delta W\): Change in weights
    \item \(W_T\): Old weights
    \item \(W_{T+1}\): New weights
\end{itemize}

The correlation matrix that drives this process is computed as the outer product of pre-synaptic and post-synaptic neuron activations, capturing how strongly neurons co-activate during behavior. In our implementation, we set the trace update factor to 0.05, meaning that 5\% of the previous trace information is replaced with new correlation data in each update. The remaining 95\% of trace information is preserved through the decay factor, creating a temporally extended memory of consistent activation patterns.

To further enhance stability and prevent runaway growth, we implement weight clipping after each update:
\begin{equation}
W_{T+1} = \text{clip}(W_{T+1}, -W_{\text{max}}, W_{\text{max}})
\end{equation}
where:
\begin{itemize}
\item \(W_{\text{max}} = 2.0\) in our implementation. 
\end{itemize}

This ensures that weights remain within biologically plausible bounds, further preventing potential explosion or saturation of connection strengths. 

This entire process creates an online adaptation method, where the robot begins each trial (lifetime) with its original evolved genome (genetic memory) and adapts on the fly according to environmental conditions and its own performance. Unlike traditional learning approaches that require extensive training on specific scenarios, our hybrid evolutionary-Hebbian system leverages a pre-optimized baseline controller while still providing a temporary online adaptation to novel or changing environments \cite{Hebbian_yuille_kersten}.

\section{Experimentation}
The experimentation we provide focuses more towards the adaptability of the robot in terms of the success rate and fitness.

\subsection{Light Sensitive}

\begin{figure}[ht]
  \centering
  \includegraphics[width=0.8\textwidth]{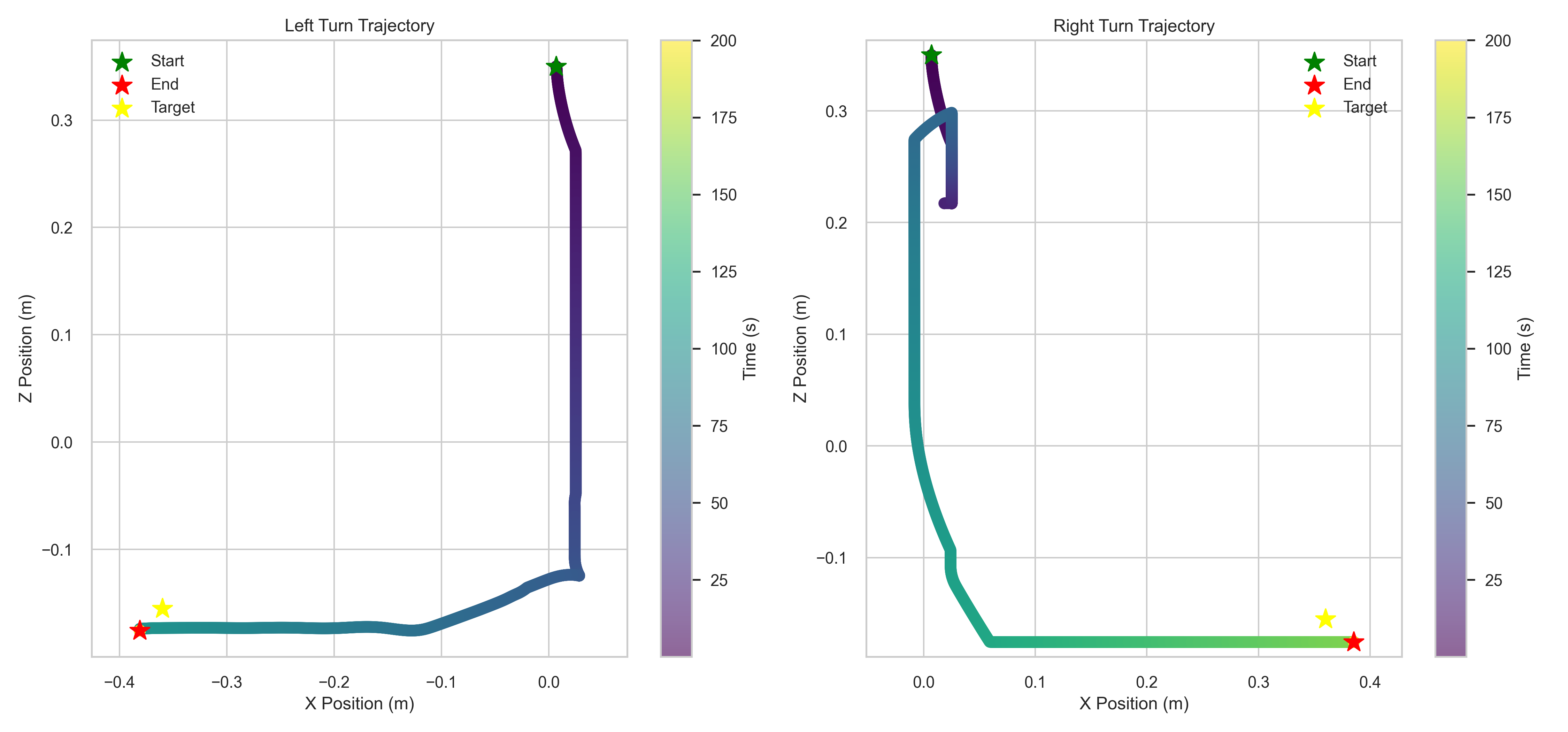}
  \caption{GA trajectory in high luminosity}
  \label{fig:initial_high_traj}
\end{figure}

\begin{figure}[ht]
  \centering
  \includegraphics[width=0.8\textwidth]{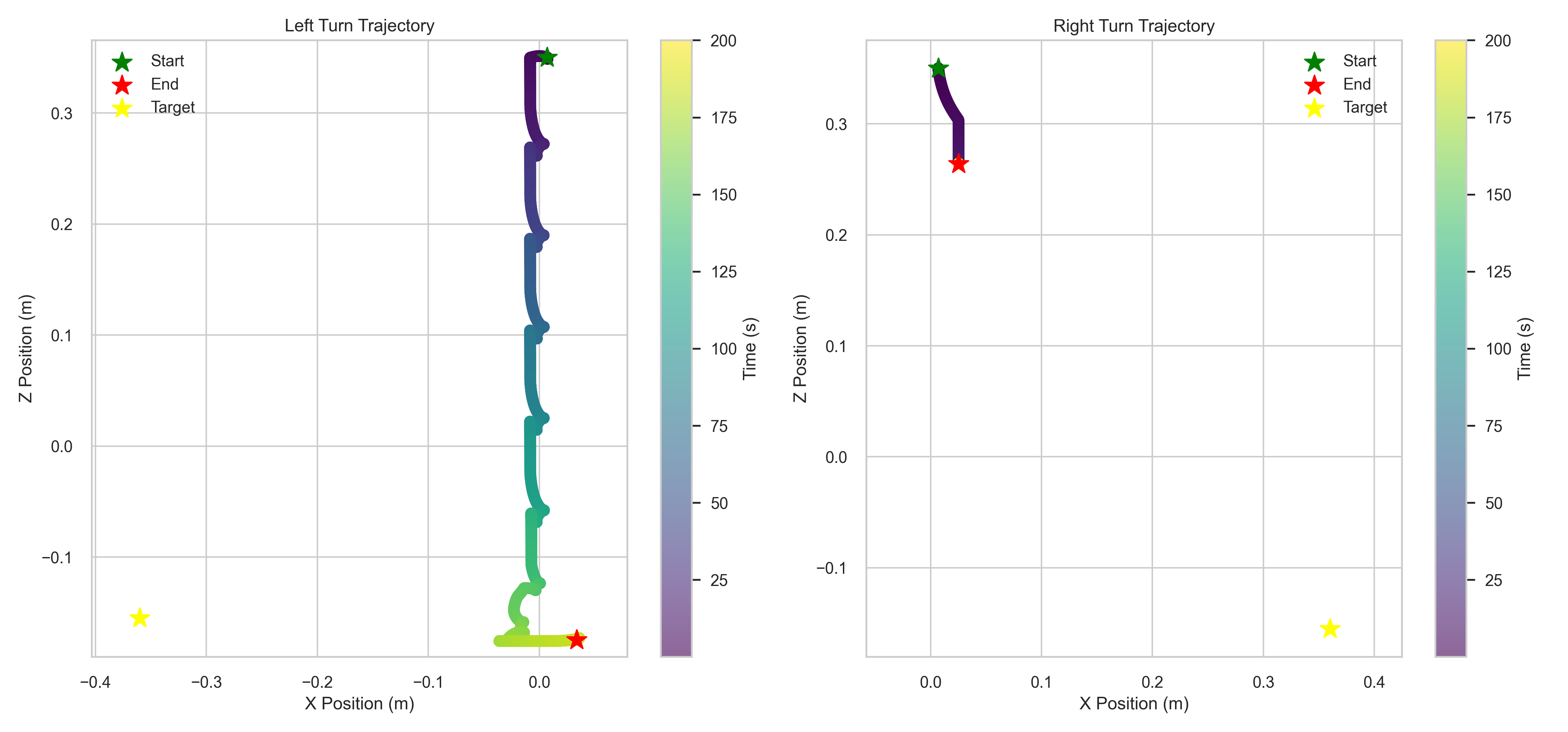}
  \caption{GA trajectory in low luminosity}
  \label{fig:initial_low_traj}
\end{figure}

\begin{figure}[ht]
  \centering
  \includegraphics[width=0.8\textwidth]{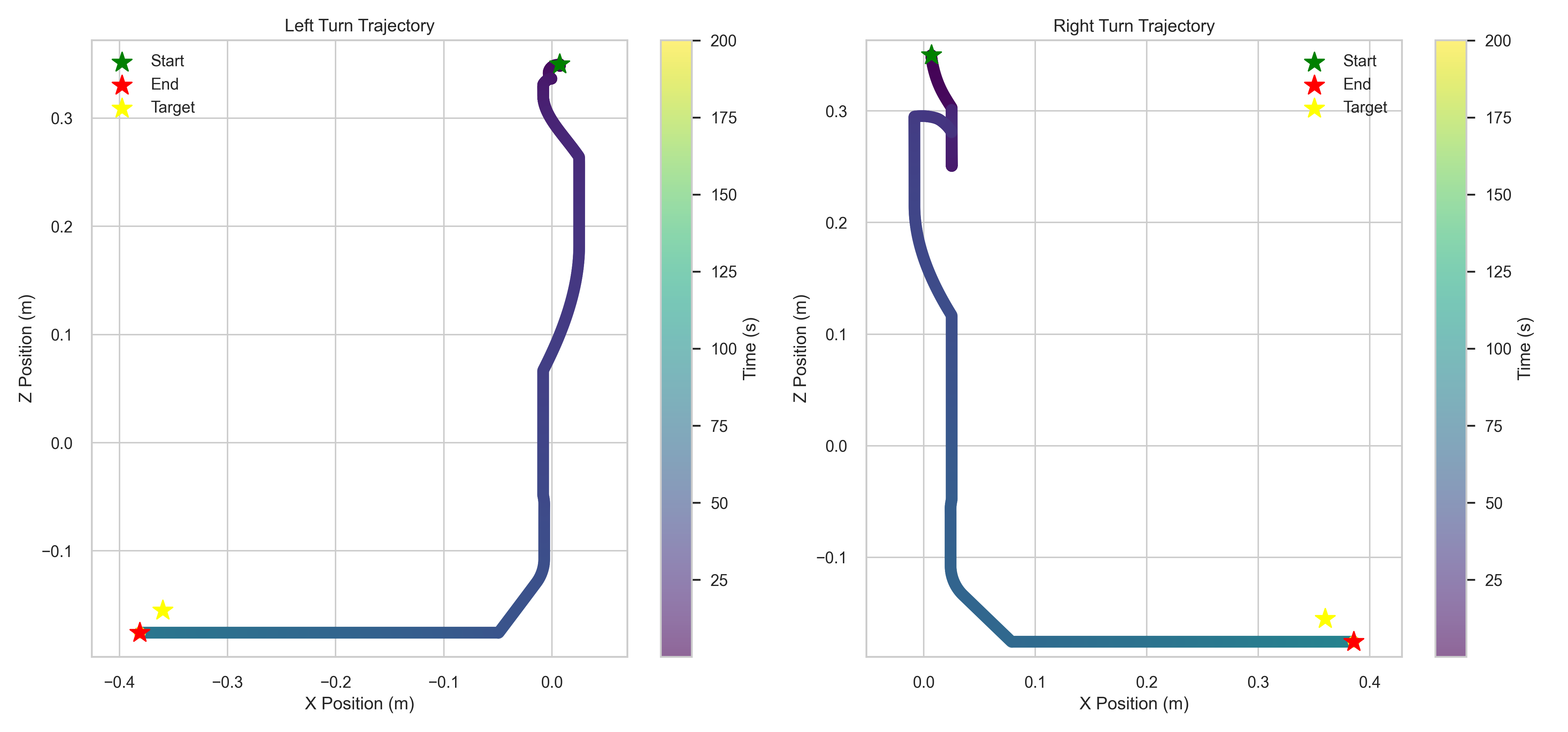}
  \caption{Synaptic GA trajectory in low luminosity}
  \label{fig:synaptic_low_traj}
\end{figure}

\begin{figure}[ht]
  \centering
  \includegraphics[width=0.8\textwidth]{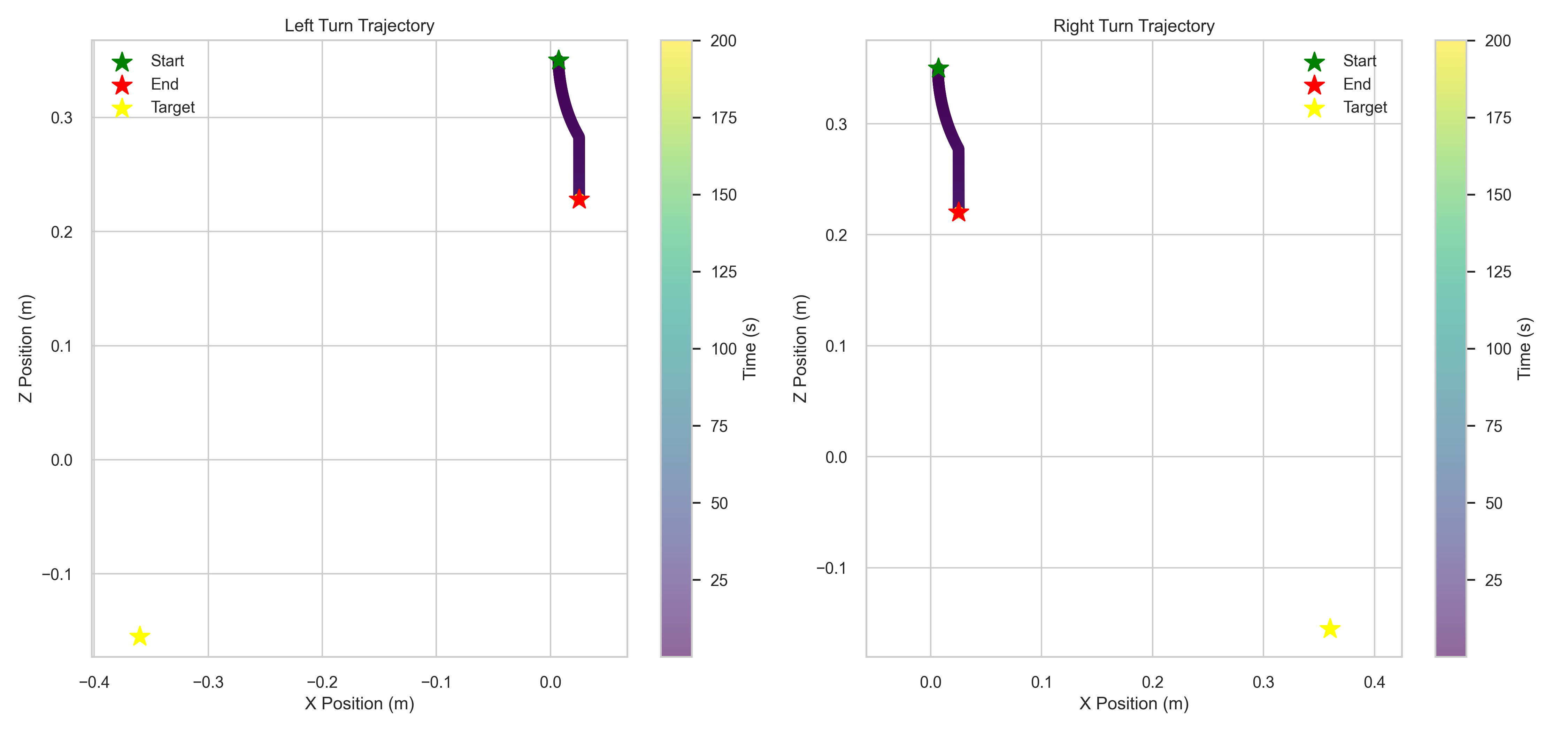}
  \caption{Synaptic GA trajectory in high luminosity}
  \label{fig:synaptic_high_traj}
\end{figure}

\textbf{\textit{Note:} Figures:\ref{fig:initial_low_traj}, \ref{fig:initial_high_traj}, \ref{fig:synaptic_low_traj}, and \ref{fig:synaptic_high_traj} are plotted upside down in correlation with the actual environment.}

The light-sensitive experiment examines the effect of light luminosity on the gene's performance. The initial gene was trained in a high-luminosity environment with a luminosity value of 1.0, using the TexturedBackgroundLight configuration, which controls our environment's light. \textbf{Fig:\ref{fig:initial_high_traj}} demonstrates the robot trajectory from start to goal position. For this, Webots' built-in positions are used to demonstrate, as light changes in the environment cause inconsistency with the tracker.

The figure above shows the expected behavior in the environment it was trained on. We then apply an environmental change in the luminosity from 1.0 to 0.1, as shown in \textbf{Fig: \ref{fig:initial_low_traj}}. The robot's trajectory became unbalanced due to the unfamiliar light conditions, as light sensors input to the network and they received unfamiliar inputs causing incorrect movements. The changes can be applied in the initial training phase to increase the gene's adaptability, but in that case, training time increases.

In the demonstration in the previous figures, we use the first phase of the controller trained only on GA. In the base environment, the success rate and the position error were higher with values of 0.0295 and 0.0327, while in the changed light condition, the position error was at 0.3935 and 0.5464. The position error shown at the start is only due to the robot hitting dead end, but it has passed next to the estimated goal position.

Shown in Figure \textbf{Fig:\ref{fig:synaptic_low_traj}} is the Hebbian adaptation to the light change.

\begin{figure}[ht]
  \centering
  \includegraphics[width=0.8\textwidth]{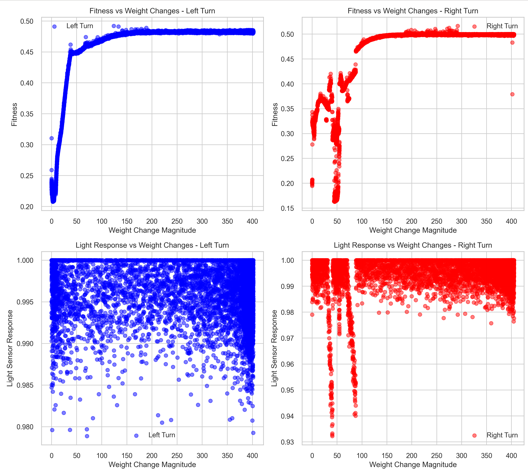}
  \caption{Weight light correlation}
  \label{fig:weight_corr}
\end{figure}

While the paths seem less stable compared to the GA, the success rate was determined through getting the same position error. Further analysis of weight changes correlations and overall changes is shown in the correlation plot \textbf{Fig:\ref{fig:weight_corr}}], which illustrates the correlation between weight changes and fitness levels in the table above, and then in correlation to light sensor readings in button with left turn plotted in blue and right turn plotted in red. Minor adjustments occur at lower fitness levels, and more stable changes are observed at higher fitness levels due to our fitness scale. While not in direct correlation, the light response also causes changes in the weights, which are more spread in the trial that has no light source.

With Hebbian learning, we achieved better success rates across environments. The behavior in low-light environments was observed, and this analysis highlights the potential of Hebbian learning to refine robot behavior and enhance adaptability.

\textbf{Table:\ref{tab:runtime}} provides a summary of the runtime evaluations for each gene.

\begin{table}[!ht]
\centering
\resizebox{0.5\textwidth}{!}{%
\begin{tabular}{|l|c|c|c|c|c|c|}
\hline
 & \multicolumn{2}{c|}{\textbf{Low Light GA}} & \multicolumn{2}{c|}{\textbf{High Light GA}} & \multicolumn{2}{c|}{\textbf{Low Light Hebbian}} \\
 & \textbf{Left} & \textbf{Right} & \textbf{Left} & \textbf{Right} & \textbf{Left} & \textbf{Right} \\
\hline
\textbf{Distance} & 0.7933 & 0.0969 & 0.9050 & 1.0699 & 0.9063 & 1.1082 \\
\hline
\textbf{Average Speed} & 0.0040 & 0.0005 & 0.0045 & 0.0053 & 0.0045 & 0.0055 \\
\hline
\textbf{Position Error} & 0.3935 & 0.5364 & 0.0295 & 0.0327 & 0.0295 & 0.0327 \\
\hline
\textbf{Average Weight Change} & \multirow{2}{*}{NA} & \multirow{2}{*}{NA} & \multirow{2}{*}{NA} & \multirow{2}{*}{NA} & \multirow{2}{*}{282.6898} & \multirow{2}{*}{259.8504} \\
 & & & & & & \\
\hline
\end{tabular}%
}
\caption{Runtime evaluations: Experiment 1}
\label{tab:runtime}
\end{table}

Applying Hebbian learning in a high-light environment yields limited benefits, as adaptation is unnecessary under conditions the robot is already familiar with. As shown in \textbf{Fig:\ref{fig:synaptic_high_traj}}, the robot fails to perform the task correctly, indicating that no adaptation is needed in this scenario. Currently, this is a limitation in the methodology, and a necessary improvement is needed where determining a change of environment will cause a higher base Hebbian rate for scale. Smaller changes will cause smaller rates, and no changes should yield a rate of zero. Currently, this method is limited to parameter tuning; in this experiment, our base rate is 0.002, which in this case is a high rate.

The reason the adaptation here causes the failure is due to having a high rate. There would be a large correlation between the light sensors and moving left since the background light is already very high. The robot just headed left straight away and was stuck near the wall.

\subsection{Obstacle Avoidance}
\begin{figure}[ht]
  \centering
  \includegraphics[width=0.8\textwidth]{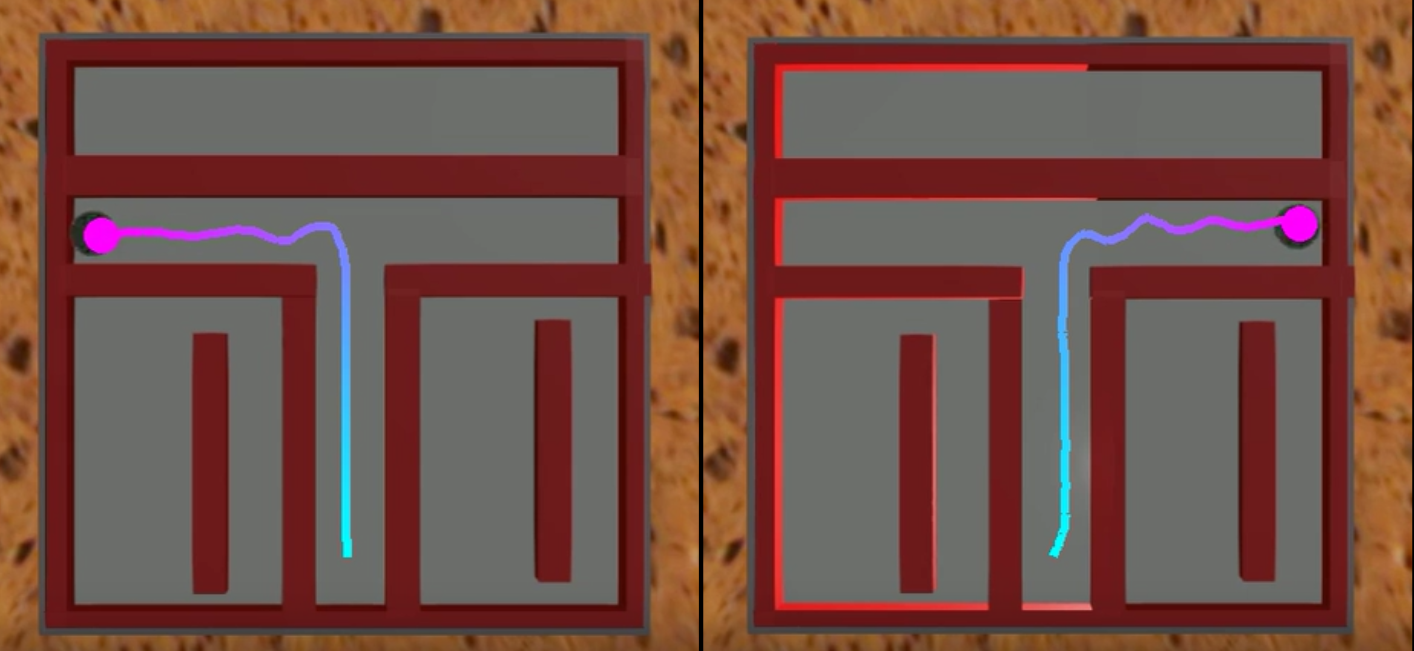}
  \caption{Expected trajectory}
  \label{fig:co0_traj}
\end{figure}

In the next experiment, we examine a gene that evolved to avoid walls and align with corridors. To analyze the robot's trajectory, we use the CoTracker by Meta \cite{karaev23cotracker} to track its movements in video recordings. This allows us to observe the trajectory in the actual environment and detect any collisions. First, in the trained environment, as shown in \textbf{Fig:\ref{fig:co0_traj}}, we observe the expected trajectory in both runs.

\begin{figure}[ht]
  \centering
  \includegraphics[width=0.8\textwidth]{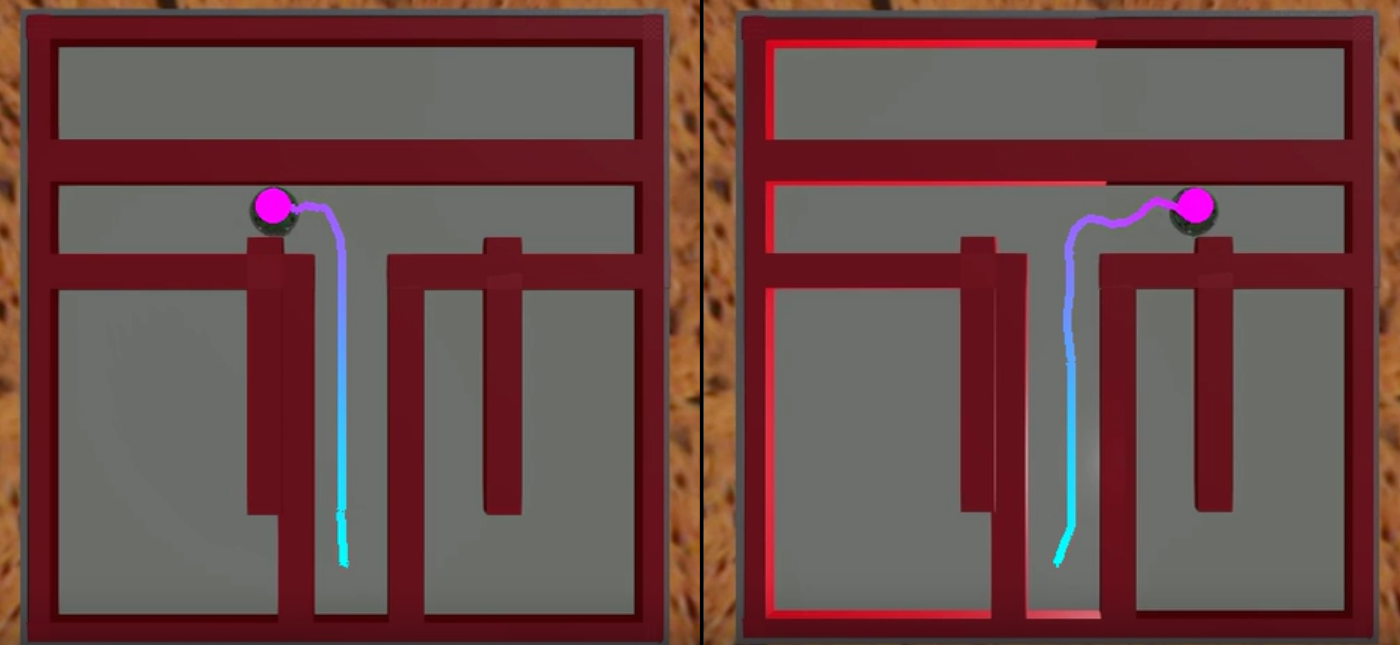}
  \caption{trajectory failure with obstacle}
  \label{fig:co1_traj}
\end{figure}

\begin{figure}[ht]
  \centering
  \includegraphics[width=0.8\textwidth]{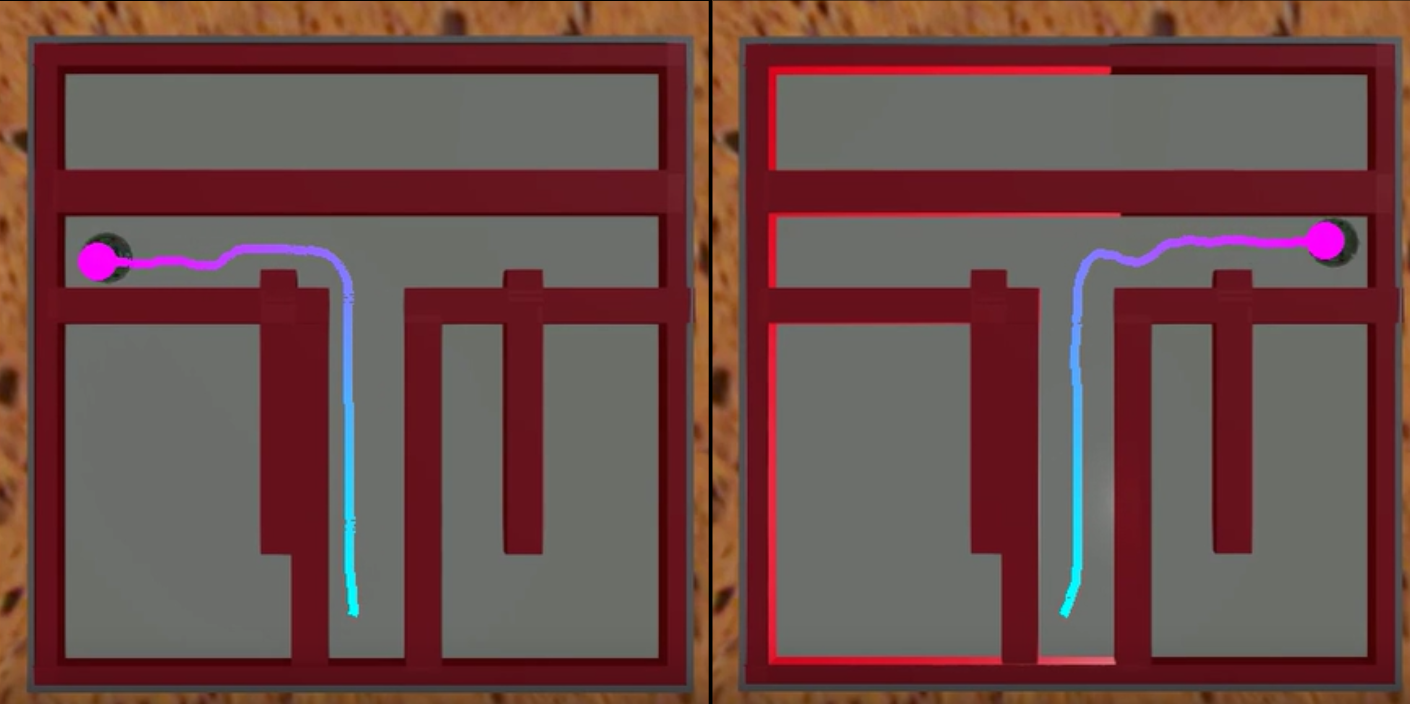}
  \caption{Hebbian adaption}
  \label{fig:co2_traj}
\end{figure}

\begin{figure}[ht]
  \centering
  \includegraphics[width=0.8\textwidth]{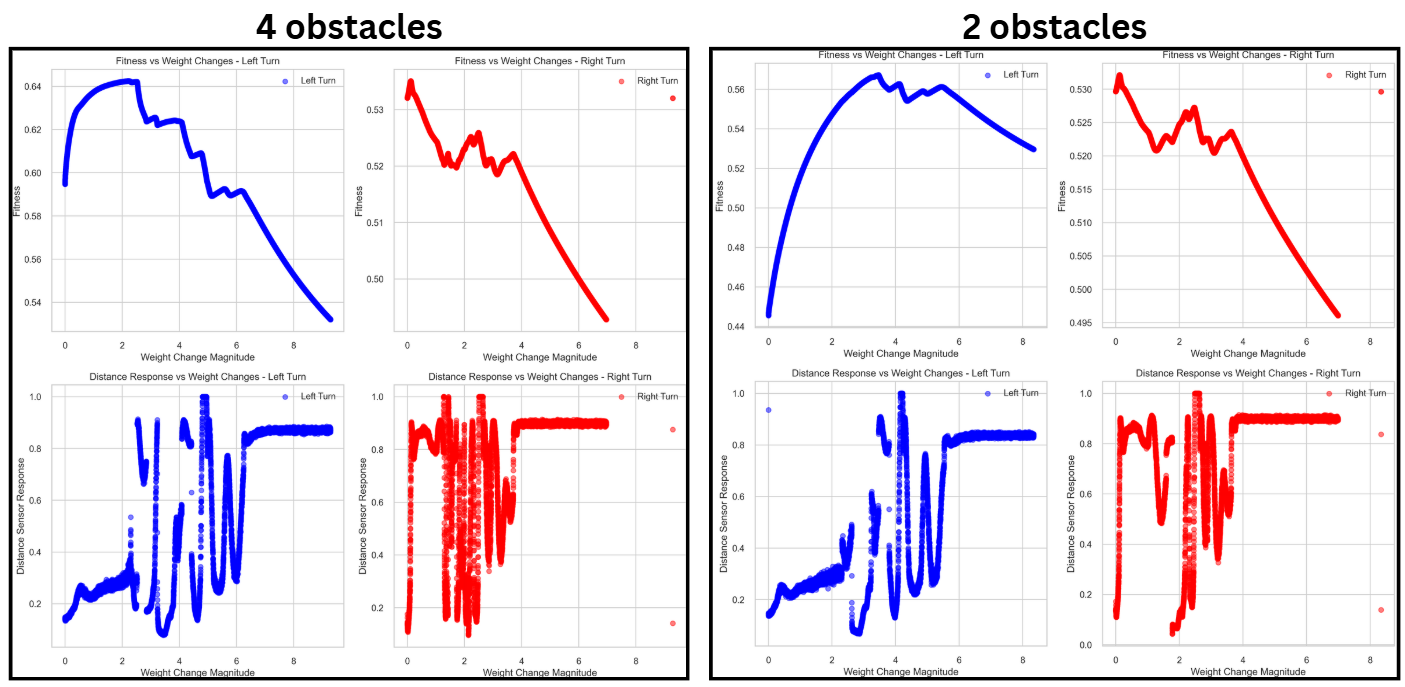}
  \caption{Weight distance correlation}
  \label{fig:syn_dif}
\end{figure}

2 obstacles are added to tighten the corridors at the junction turn, in \textbf{Fig:\ref{fig:co1_traj}} shows that the robot crashes into them if we run GA alone. The task itself is not impossible robot can avoid them but struggles with the changes due to unfamiliarity with the policy it has learned. The original genotype does not react quickly enough. Retraining the gene would not be ideal as GA training takes time.
Hebbian learning provides a quick adaptation method as mention before once a learning rate is determined. In a previous experiment, a learning rate of 0.002 was used, which, in the context of our experimentation, was considered very high. In this experiment, a lower learning rate of 0.000015 was applied, as shown in \textbf{Fig:\ref{fig:co2_traj}} where the robot quickly adapted to the new environment and successfully avoided the new environmental changes and reached its goal. 
Two additional obstacles were introduced into the environment. The correlation plot in \textbf{Fig:\ref{fig:syn_dif}} same as before but table below shows the weight change against the distance sensor response not light sensor response, shows two different runs with the same code but different environments: one with two obstacles and the other with four obstacles. Changes in the plot indicate reactive behavior even tho no code or parameter changes were applied. 

In total, for this experiment, we ran six runs across the base environment, two added obstacles, and four added obstacles, each evaluated with GA or Hebbian learning. Statistics of each experiment are shown in \textbf{Tables:\ref{tab:runtime2_GA} and \ref{tab:runtime2_heb}}. \\

\begin{table}[!ht]
\centering
\resizebox{0.4\textwidth}{!}{%
\begin{tabular}{|l|c|c|c|c|c|c|}
\hline
 & \multicolumn{2}{c|}{\textbf{base GA}} & \multicolumn{2}{c|}{\textbf{2 Obstacles GA}} & \multicolumn{2}{c|}{\textbf{4 Obstacles GA}} \\
 & \textbf{Left} & \textbf{Right} & \textbf{Left} & \textbf{Right} & \textbf{Left} & \textbf{Right} \\
\hline
\textbf{Success (0/1)} & 1 & 1 & 0 & 0 & 0 & 0 \\
\hline
\textbf{Time-to-Goal (s)} & 213.82 & 174.62 & nan & nan & nan & nan \\
\hline
\textbf{Path Length (m)} & 0.9114 & 0.8854 & 0.6195 & 0.7298 & 0.6190 & 0.5520 \\
\hline
\textbf{Average Speed (m/s)} & 0.0036 & 0.0035 & 0.0025 & 0.0029 & 0.0025 & 0.0022 \\
\hline
\textbf{Final Position Error} & 0.0221 & 0.0319 & 0.2592 & 0.1225 & 0.2592 & 0.3428 \\
\hline
\end{tabular}%
}
\caption{Runtime evaluations: Experiment 2 GA}
\label{tab:runtime2_GA}
\end{table}

\vspace*{-10pt}

\begin{table}[!ht]
\centering
\resizebox{0.4\textwidth}{!}{%
\begin{tabular}{|l|c|c|c|c|c|c|}
\hline
 & \multicolumn{2}{c|}{\textbf{base Hebbian}} & \multicolumn{2}{c|}{\textbf{2 Obstacles Hebbian}} & \multicolumn{2}{c|}{\textbf{4 Obstacles Hebbian}} \\
 & \textbf{Left} & \textbf{Right} & \textbf{Left} & \textbf{Right} & \textbf{Left} & \textbf{Right} \\
\hline
\textbf{Success (0/1)} & 1 & 1 & 1 & 1 & 1 & 1 \\
\hline
\textbf{Time-to-Goal (s)} & 192.32 & 156.77 & 192.29 & 156.48 & 191.26 & 161.25 \\
\hline
\textbf{Path Length (m)} & 0.8915 & 0.8828 & 0.8915 & 0.8836 & 0.8893 & 0.8813 \\
\hline
\textbf{Average Speed (m/s)} & 0.0036 & 0.0035 & 0.0036 & 0.0035 & 0.0036 & 0.0035 \\
\hline
\textbf{Final Position Error} & 0.0233 & 0.0327 & 0.0232 & 0.0327 & 0.0233 & 0.0327 \\
\hline
\textbf{Weight Change Mean} & \multirow{2}{*}{2.9702} & \multirow{2}{*}{2.5906} & \multirow{2}{*}{3.0132} & \multirow{2}{*}{2.6165} & \multirow{2}{*}{3.1106} & \multirow{2}{*}{2.6973} \\
 & & & & & & \\
\hline
\end{tabular}%
}
\caption{Runtime evaluations: Experiment 2 Hebbian}
\label{tab:runtime2_heb}
\end{table}

In terms of success rate, Hebbian learning in this experiment achieved a 100\% success rate across all environments. Increasing weight change averages with more obstacles indicating GA attained 1 success only in the Base environment but failed in both two- and four-obstacle environments, the success with Hebbian was done with no extra training.
\section{Benefits and Limitations}

This approach offers several benefits. First, it leverages a pre-evolved policy to ensure competent baseline behavior, while online Hebbian adaptation refines performance as needed rather than learning from scratch. The robot’s core functionality is preserved, with plasticity serving only as a temporary overlay.
as demonstrated in the experimentation, we avoid catastrophic interference: after adaptation, the network returns to its original weights, so it doesn’t accumulate unnecessary changes during a second runtime, the adaptation is fast, happening within the runtime, which is crucial for real-world scenarios where a robot cannot undergo lengthy retraining upon encountering a new obstacles. 

However, a significant limitation in our current implementation, is the use of static Hebbian Learning parameter, or the maximum learning rate in this case, which are not adjusted based on the magnitude of environmental changes. In our runs, we found that if the environment change is minor, a large $N$ can cause the controller to overreact causing unstable behavior and degraded performance. On the other hand major and more persistent changes such as lighting conditions demand higher $N$ for quick adaptation and a too-small $N$ means the robot adapts too slowly within the given trial, the system would ideally need to detect the degree of novelty or performance drop and scale its $N$ accordingly and even turn off plasticity when no changes are detected and normal routine is activated. Despite these limitations, our results demonstrated a clear improvement in adaptability as experiment has shown. 
\section{Discussion}
Our experiments propose that integrating evolutionary optimization with Hebbian plasticity can produce a controller capable of remarkable adaptability to novel environments without requiring further training. This approach fills a gap in evolutionary robotics, the absence of online adaptation in conventional GA-evolved controllers.
\begin{itemize}

\item \textbf{How does this work fill the gap?}: GA produce a fixed policy encoded in the genome. If it encounters environmental changes, that fixed policy may no longer be appropriate. The GA would typically have to evolve again. This approach instead keeps the genome fixed as genetic memory and adds a mechanism for the phenotype to adjust itself, going back to the Baldwin effect. This can be seen as an illustration of the benefit of lifetime learning. In fact, we prove this through our statistical results in \textbf{Table: \ref{tab:runtime2_heb}} at the "Time-to-Goal (s)" During "Left" turns. In the base environment, the robot reached the goal in 192.32 seconds. Adding 2 obstacles made it slightly faster at 192.29 seconds. 4 obstacles cut the time by a whole second at 191.26 seconds. Although this is not active in all cases, as in the left turn, 4 obstacles was slower by more than 5 seconds. However, 2 obstacles was still faster than the base. This still aligns with the theory that individuals (robots) that can adapt during their lifetime achieve higher fitness in changed environments than those that cannot. At the same time, individuals that adapt more effectively can outperform others under certain conditions. The adaptability is illustrated as well in Weight Change Mean, where more obstacles required more weight changes (adaptability).

\item \textbf{Biological inspiration}: The inspiration we mentioned earlier comes from separating learning and memory. The genotype (memory) remains unchanged by learning towards the end of the lifetime, similar to that of animals where their DNA stays intact even when they have to adapt. Their genetic memory stays the same, and the learning improves the animal's performance, which indirectly influences evolutionary success. This directly proves the biological inspiration towards the approach.

\item \textbf{Neuromodulation}: Our approach uses the existing fitness function as the learning signal, forming a neuromodulated plasticity. The "neuromodulator" is the reward signal (fitness). Imagine a modulatory neuron releasing a chemical proportional to the current fitness. This idea, demonstrated by \cite{soltoggio2008evolutionary}, showed how a modulatory signal enables an effective reward-based learning approach, while a simpler implementation directly aligns with the principles. 
We propose that reward feedback can be a powerful driver for rapid learning.

\item \textbf{Responsive adaptation}: The Hebbian experiments demonstrate that, without additional training, the robot can rapidly adapt to its surroundings. Our algorithm's Hebbian component acts more like a reflex for the robot than a learning algorithm in the traditional sense. It's as if the robot is reasoning about its situation. Once the situation is over, the gene returns to its prior state. In \textbf{Fig:\ref{fig}}, we demonstrate how the Hebbian mechanism causes a reaction when 2 of the robot's distance sensors (red neurons in the first layer) are activated, causing a correlation change (yellow connections) to neurons on the second layer.
We propose that Hebbian learning can serve as a direct feedback mechanism as a form of real-time “thinking” into controllers. 
\end{itemize}
\begin{figure}[ht]
\centering
\includegraphics[width=0.8\textwidth]{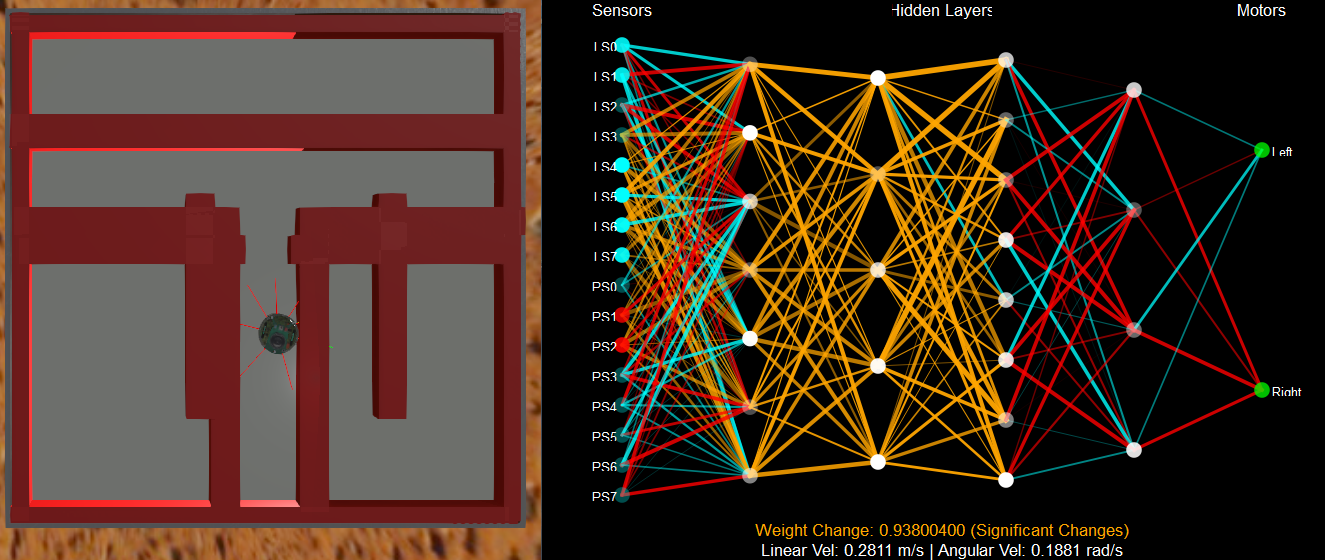}
\caption{Neural Network Weight Changes}
\label{fig:network}
\end{figure}
\section{Future Work}
\begin{itemize}
    \item \textbf{Plasticity}\\
    Due to current limitations, we plan to implement a mechanism to detect environmental changes or context shifts. given example by streaming sensory data through a spike‑dependent network to alternate from the base Hebbian rate to adaptive Hebbian rates based on memory sensor shift.
     
    \item \textbf{Network}\\
    Experiments used a relatively simple MLP with low-dimensional sensory input to benchmark the plasticity operator. future implementations aim to test more complex sensory data, for example point clouds or LiDAR—while retaining the same plasticity mechanism.

    \item \textbf{Multiple robots and tasks}\\
    All current experiments use a single e‑puck in a T‑maze. approach’s generality depends on porting the genotype–plasticity split to to different robot platforms, drones, or legged robots used by \cite{10.1145/3449639.3459317} and \cite{10.1145/3638529.3654011} to further test adaptation, test in different and more complex terrains.
\end{itemize}

\section{Conclusion}
Our approach demonstrates a valuable split between learned memory and direct adaptation. By outlining existing algorithms but redefining their architecture, we have proposed a novel methodology. While it comes with limitations, future work aims to address them to truly develop a controller capable of reliable zero-shot adjustments across generalized terrains and platforms.

\footnotesize
\bibliographystyle{apalike}
\bibliography{main}

\begin{thebibliography}{}

\bibitem[Caporale and Dan, 2008]{Caporale2008}
Caporale, N. and Dan, Y. (2008).
\newblock Spike timing–dependent plasticity: A hebbian learning rule.
\newblock {\em Annual Review of Neuroscience}, 31:25--46.

\bibitem[Chiang, 2010]{5533882}
Chiang, C.-H. (2010).
\newblock A genetic programming based rule generation approach for intelligent control systems.
\newblock In {\em 2010 International Symposium on Computer, Communication, Control and Automation (3CA)}, volume~1, pages 104--107.

\bibitem[Ferigo et~al., 2024]{10.1145/3638529.3654011}
Ferigo, A., Cunegatti, E., and Iacca, G. (2024).
\newblock Neuron-centric hebbian learning.
\newblock In {\em Proceedings of the Genetic and Evolutionary Computation Conference}, GECCO '24, page 87–95, New York, NY, USA. Association for Computing Machinery.

\bibitem[Hebb, 1949]{Hebb1949}
Hebb, D. (1949).
\newblock {\em The Organization of Behavior: A Neuropsychological Theory}.
\newblock John Wiley \& Sons, New York.

\bibitem[Karaev et~al., 2024]{karaev23cotracker}
Karaev, N., Rocco, I., Graham, B., Neverova, N., Vedaldi, A., and Rupprecht, C. (2024).
\newblock Cotracker: It is better to track together.
\newblock In {\em Proc. {ECCV}}.

\bibitem[Masumori et~al., 2024]{10.1162/isal_a_00799}
Masumori, A., Doi, I., and Ikegami, T. (2024).
\newblock Plasticity in swarming behavior: Introducing social network to boids model.
\newblock In {\em Proceedings of the 2024 Artificial Life Conference (ALIFE 2024)}, Artificial Life Conference Proceedings, page 105.

\bibitem[Mori and Kita, 2000]{mori2000_ga_dynamic}
Mori, N. and Kita, H. (2000).
\newblock Genetic algorithms for adaptation to dynamic environments -- a survey.
\newblock In {\em Proceedings of the 26\textsuperscript{th} Annual Conference of the IEEE Industrial Electronics Society (IECON 2000)}, pages 2947--2952, Nagoya, Japan. IEEE.

\bibitem[Najarro and Risi, 2020]{NEURIPS2020_ee23e7ad}
Najarro, E. and Risi, S. (2020).
\newblock Meta-learning through hebbian plasticity in random networks.
\newblock In Larochelle, H., Ranzato, M., Hadsell, R., Balcan, M., and Lin, H., editors, {\em Advances in Neural Information Processing Systems}, volume~33, pages 20719--20731. Curran Associates, Inc.

\bibitem[Pedersen and Risi, 2021]{10.1145/3449639.3459317}
Pedersen, J.~W. and Risi, S. (2021).
\newblock Evolving and merging hebbian learning rules: increasing generalization by decreasing the number of rules.
\newblock In {\em Proceedings of the Genetic and Evolutionary Computation Conference}, GECCO '21, page 892–900, New York, NY, USA. Association for Computing Machinery.

\bibitem[Soltoggio et~al., 2008]{soltoggio2008evolutionary}
Soltoggio, A., Bullinaria, J.~A., Mattiussi, C., D{\"u}rr, P., and Floreano, D. (2008).
\newblock Evolutionary advantages of neuromodulated plasticity in dynamic, reward-based scenarios.
\newblock In {\em Proceedings of the 11th international conference on artificial life (Alife XI)}, page 569. MIT Press.

\bibitem[Sznajder et~al., 2012]{sznajder2012adaptive}
Sznajder, B., Sabelis, M.~W., and Egas, M. (2012).
\newblock How adaptive learning affects evolution: reviewing theory on the baldwin effect.
\newblock {\em Evolutionary biology}, 39:301--310.

\bibitem[Yaman et~al., 2021]{Yaman_2021}
Yaman, A., Iacca, G., Mocanu, D.~C., Coler, M., Fletcher, G., and Pechenizkiy, M. (2021).
\newblock Evolving plasticity for autonomous learning under changing environmental conditions.
\newblock {\em Evolutionary Computation}, 29(3):391–414.

\bibitem[Yuille and Kersten, 2020]{Hebbian_yuille_kersten}
Yuille, A. and Kersten, D. (2020).
\newblock Hebbian learning and probabilistic models of visual cognition.
\newblock In {\em Lecture Notes on Probabilistic Models of Visual Cognition}.

\bibitem[Zenke et~al., 2017]{ZENKE2017166}
Zenke, F., Gerstner, W., and Ganguli, S. (2017).
\newblock The temporal paradox of hebbian learning and homeostatic plasticity.
\newblock {\em Current Opinion in Neurobiology}, 43:166--176.
\newblock Neurobiology of Learning and Plasticity.

\end{thebibliography}
\end{document}